\documentclass[letterpaper, 10 pt, conference]{ieeeconf}
\IEEEoverridecommandlockouts                              

\usepackage{cite}
\usepackage{multirow}
\usepackage{algorithm}  	
\usepackage{algorithmic}
\usepackage{graphicx}
\usepackage{times}
\usepackage{amsmath}
\usepackage{amssymb}
\usepackage{subfigure}
\usepackage{graphics} 
\graphicspath{ {./imgs/} }

\usepackage[utf8]{inputenc}
\usepackage{boldline}
\usepackage{multicol}
\usepackage[pdfstartview=FitH,bookmarksnumbered=true,colorlinks,bookmarksopen=true]{hyperref}

\title{\LARGE \bf
Adversarial Skill Learning for Robust Manipulation
}

\author{Pingcheng Jian$^{\dagger}$, Chao Yang$^{\dagger}$, Di Guo, Huaping Liu$^{*}$, Fuchun Sun
\thanks{$\dagger$These two authors contributed equally.}
\thanks{The authors are with the Beijing National Research Center for Information Science and Technology (BNRist), State Key Lab on Intelligent Technology and Systems, Department of Computer Science and Technology, Tsinghua University.}
\thanks{$^{*}$Corresponding author to provide e-mail: hpliu@tsinghua.edu.cn}%
}
\begin{document}

\maketitle
\thispagestyle{empty}
\pagestyle{empty}

\begin{abstract}

Deep reinforcement learning has made significant progress in robotic manipulation tasks and it works well in the ideal disturbance-free environment. However, in a real-world environment, both internal and external disturbances are inevitable, thus the performance of the trained policy will dramatically drop. To improve the robustness of the policy, we introduce the adversarial training mechanism to the robotic manipulation tasks in this paper, and an adversarial skill learning algorithm based on soft actor-critic (SAC) is proposed for robust manipulation. Extensive experiments are conducted to demonstrate that the learned policy is robust to internal and external disturbances. Additionally, the proposed algorithm is evaluated in both the simulation environment and on the real robotic platform. 
\end{abstract}

\section{INTRODUCTION}
In recent years, policies trained by reinforcement learning (RL) methods~\cite{gullapalli1994acquiring, asada1996purposive} have been widely used in robot control and have achieved great success in lots of scenarios. However, ordinary RL policies are usually fragile and can easily fail when disturbances occur. In a real-world environment, both internal and external disturbances inevitably exist, thus it is crucial for a practical robot to be robust enough to adapt to them automatically. 

In fact, robust control of robot has been a heated research topic for decades~\cite{petersen2012robust, slotine1985robust, bowman2019intent}. With the application of reinforcement learning methods in robot control, researchers have developed several methods to improve the robustness of reinforcement learning~\cite{morimoto2005robust, pinto2017robust}. Among them, adversarial reinforcement learning is a new idea developed in recent years~\cite{gu2019adversary, pinto2017supervision, wulfmeier2018incremental, porav2018adversarial, ghosh2018verifying}. 

Actually, the adversarial method is first proposed in supervised learning and applied in the field of image classification. The adversarial samples are introduced in order to train a more robust classifier ~\cite{goodfellow2015explaining, brendel2018decision}. Some other works~\cite{miyato2016adversarial, madry2018towards, shafahi2019adversarial,zhang2019you, zhu2019freelb} also try to use the adversarial training mechanism to improve the robustness of neural network against attack and disturbance. Tony et al.~\cite{chen2019adversarial} make a comprehensive survey on adversarial attacks from the perspective of AI security and briefly introduced the defense approaches against different types of adversarial attacks. In the Robust Adversarial Reinforcement Learning (RARL)~\cite{pinto2017robust}, the adversarial training method is introduced into the field of reinforcement learning. They use a protagonist agent and an adversary agent to confront each other and optimize alternately. Adversarial training enables the protagonist agent to experience more extreme situations, making it more robust after training. The RARL methods have been applied in both the simulation and real-world environments, demonstrating the robustness of the trained agent ~\cite{gu2019adversary, pinto2017supervision}. Huang et al.~\cite{huang2017adversarial} evaluate the robustness of learned policy through an FGSM based attack on Atari games with discrete actions. Gu et al. ~\cite{gu2019adversary} develop an algorithm called adversary robust A3C to improve the robustness of the baseline A3C algorithm using adversary training, and it is evaluated on a pendulum hardware platform.

\begin{figure}[t!]
    \centering
    \includegraphics[width=0.45\textwidth]{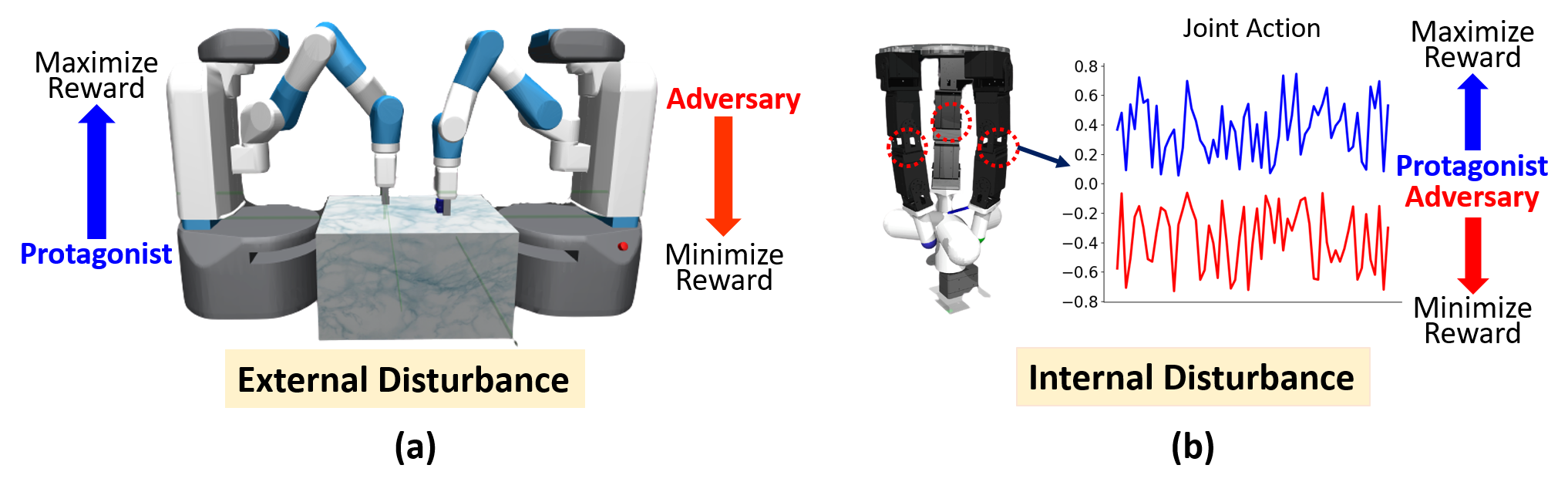}
    \caption{Adversarial skill learning for robust manipulation. The protagonist tries to maximize the reward while the adversary tries to minimize it. (a) An external disturbance, the protagonist is disturbed by another adversary robot. (b) An internal disturbance, the protagonist joint action is under the disturbance of adversarial action on each joint of the robot.}
    \label{fig:title-figure}
\end{figure}

However, research on this subject has been mostly restricted to dense reward tasks that are not suitable for some robotics manipulation tasks with sparse rewards. Besides, to the best of our knowledge, few researchers have investigated different types of disturbance in robotic manipulation tasks comprehensively and proposed a robust RL method, which is more generalized. From our perspective, the disturbances can be divided into two classes: internal disturbance and external disturbance. We define the internal disturbance as the sum of various internal noise on the joints of a robot, making the robot unable to carry out the ideal action given by its controller. The external disturbance is executed outside the robot and can hinder the robot from accomplishing its task. It is important to analyze these two kinds of disturbance separately and develop a general method to improve the robot's robustness against both of them. The main idea of this paper is demonstrated in Fig.\ref{fig:title-figure}. For the external and internal disturbance in robotic manipulation tasks, protagonist agents try to maximize the reward while adversary agents try to minimize it. The protagonist agent is expected to be more robust after the adversarial training. 

In this paper, we develop an adversarial training method based on soft actor-critic (SAC)~\cite{haarnoja2018soft}, which can improve robot robustness against internal and external disturbances on robotic manipulation tasks. Our proposed algorithm can be applied against both the internal noise in robot joint space and the external disturbance from another adversarial robot or human being. According to our experiments, it achieves obviously higher robustness against both the internal noise and the external disturbance. Furthermore, for sparse reward tasks, the goal-condition data augmentation method HER~\cite{andrychowicz2017hindsight} has been introduced to our adversarial training process, enabling our method to work well even in sparse reward scenarios.

The contributions of our work can be summarized as follows:

\begin{enumerate}
    \item We propose a new adversarial skill learning method for robust manipulation, which can deal with both the internal disturbance in robot joint space and the external disturbance from other robots.
    \item We show that our method works well even for the sparse reward scenarios, which often occur in practical manipulation tasks.
    \item We carry out experiments on a real robot and show the practicability of our method in real-world robot application.
\end{enumerate}

\section{Background and Preliminary}

\label{sec:preliminary}
Before we illustrate our algorithm in detail, we need to make some necessary explanations for the involved basic concepts and preliminaries.
\subsection{Standard Reinforcement Learning on MDPs}
In this work we consider a standard finite-horizon, discounted, Markov Decision
Process (MDP) defined by a tuple $(\mathcal{S}, \mathcal{A}, r, \mathcal{P}, s_0, \gamma)$, where $\mathcal{S}$ and $\mathcal{A}$ are the sets of feasible state and action; $r(s,a): \mathcal{S} \times \mathcal{A} \rightarrow \mathbb{R}$ denotes the reward function on state $s$ and action $a$; $\mathcal{P}(s'|s,a): \mathcal{S} \times \mathcal{A} \times \mathcal{S} \rightarrow [0, 1]$ characterizes the dynamics of the environment and defines the transition probability to next-step state $s'$ if the agent takes action $a$ at current state $s$; $s_0$ is the distribution of initial state and $\gamma \in (0, 1)$ is the discount factor. An agent policy $\pi(a|s): \mathcal{S} \times \mathcal{A} \rightarrow [0,1]$ defines the probability of choosing action $a$ at state $s$.
The goal of a given MDP is to find a policy $\pi_{\mu}$ parametrized by ${\mu}$ that maximizes the expected cumulative discounted reward:
\begin{align}
J^{\mu} = E_{\pi_{\mu}}[\sum_{t=0}^{T} \gamma^t r(s_t, a_t)|s_0=s],
\end{align}
where the actions $a$ in the MDP are sampled via $a\sim \pi_\mu(a|s)$.

\subsection{Soft Actor-Critic}
Soft actor-critic (SAC) is an algorithm that optimizes a stochastic policy in an off-policy way, forming a bridge between stochastic policy optimization and DDPG-style~\cite{lillicrap2016continuous} approaches. 

SAC is an actor-critic technique consists of two models: Actor and Critic. The critic network is function approximation of action-state value function $Q(s, a)$ which evaluate the quality of state-action pair $(s, a)$. It uses the prior concepts of experience replay and target network for its update. The actor policy is trained to maximize a trade-off between expected cumulative discounted reward and policy entropy. In SAC, we consider an entropy augmented objective:
\begin{align}
J^{\mu} = E_{\pi_{\mu}(a|s)}[\sum_{t=0}^{T} \gamma^t r(s_t, a_t) + \alpha H(\pi_{\mu}(a|s))],
\end{align}
where the $H(\pi_{\mu}(a|s))$ is the entropy of actor policy $\pi_{\mu}(a|s)$.
Optimizing this objective function can bring maximum entropy of the policy to encourage exploration, which can accelerate learning later on. It can also prevent the policy from prematurely converging to a bad local optimum.

\subsection{Sparse Reward}
Most of the robot manipulation tasks are sparse reward reinforcement learning problems. The agent can only get a reward when it accomplishes the task. Since completing the task by random exploration is of extremely low probability, it is hard for the agent to explore and get successful experiences. To overcome the exploration difficulty in robotic manipulation tasks, we adopt a data augmentation method--Hindsight Experience Replay (HER). The key idea of HER~\cite{andrychowicz2017hindsight} is to relabel failed rollouts as successful ones. During the exploration process of a sparse reward task, the agent fails in most rollouts. To make full use of these failure data, we relabel the final state of a failed trajectory as the desired goal state so that the agent can get the same reward when it reaches this relabeled state as the original goal. Both original successful transitions $(s_t, g, a_t, r)$ and the relabeled transitions $(s_t, g', a_t, r')$ are push into the replay buffer. Then we sample from the replay buffer to optimize the agent policy.

\section{Adversarial Skill Learning}
Here we outline the adversarial skill learning in an environment with disturbance. We first introduce the min-max adversarial training mechanism to model the environment with disturbance. Through two different forms of disturbance, we then sort out their application scenes, respectively. Finally, we provide a practical implementation for our proposed method from the stochastic policy.

\subsection{Formulating Adversarial Attack}

The adversarial training setting can be expressed as a two player $\gamma$ discounted zero-sum Markov game. In this paper, we consider a robust MDP~\cite{tessler2019action} as the tuple: $(\mathcal{S}, \mathcal{\hat{A}}, \mathcal{\bar{A}}, \mathcal{P}, r,\gamma, s_0)$ where $\mathcal{\hat{A}}$ and $\mathcal{\bar{A}}$ are the continuous actions the players can take. $\mathcal{P}: \mathcal{S} \times \mathcal{\hat{A}} \times \mathcal{\bar{A}} \times \mathcal{S} \to \mathbb{R} $ is the transition probability density and $r: \mathcal{S} \times \mathcal{\hat{A}} \times \mathcal{\bar{A}} \to R$ is the reward of both players. we parameterize actor policy $\pi_{\theta}$ and adversary policy $\pi_{\phi}$ as $\theta$ and $\phi$ respectively. The optimization problem is formulated as
\begin{align}
\min_{\phi}\max_{\theta} E[\sum_{t=0}^{T} \gamma^t r(s_t, \hat{a}_t,  \bar{a}_t)|\pi_{\theta},\pi_{\phi}],
\end{align}
where $\hat{a}_t$ and $\bar{a}_t$ are sampled from $\pi_{\theta}$ and $\pi_{\phi}$.

A zero-sum two-player game can be seen as the protagonist maximizing the $\gamma$ discounted reward while the adversary minimizing it ~\cite{pinto2017robust, perolat2015approximate, littman1994markov}. The solution to a zero-sum game is a minimax search; it is a result of Nash equilibrium. In the zero-sum game setting, we call actor policy $\pi_{\theta}$ of protagonist agent that we wish to be more robust \textit{protagonist}  and the actor policy $\pi_{\phi}$ of adversary agent who tries to fail the task \textit{adversary}. And the adversary is used to simulate the possible disturbance in the environment.
\begin{figure}[t!]
    \centering
    \includegraphics[width=0.45\textwidth]{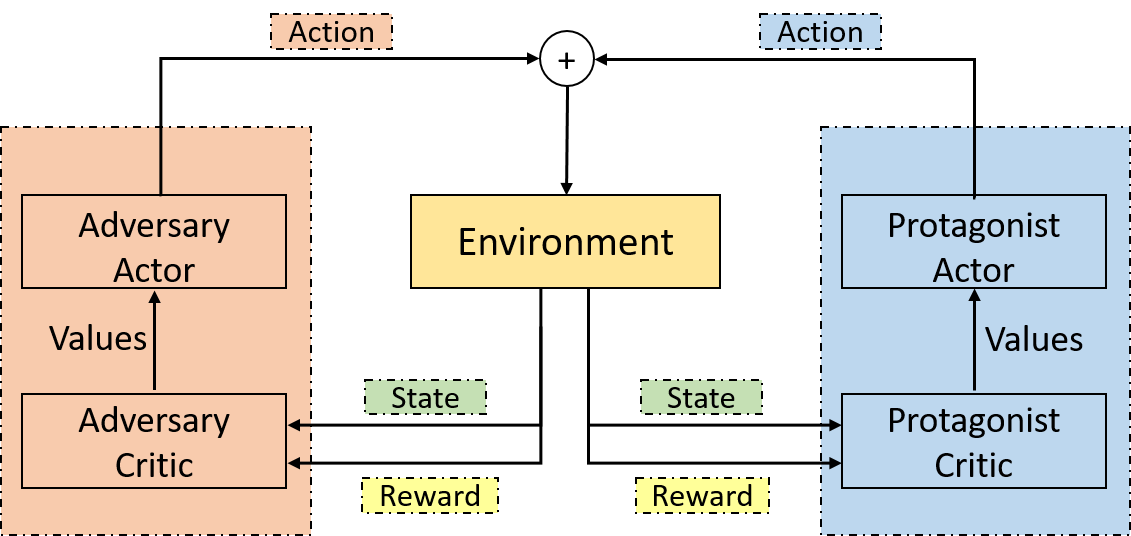}
    \caption{Adversarial attack mechanism. Both agents are based on the actor-critic structure. The protagonist outputs an action to maximize the reward and accomplish the task while the adversary outputs an action to minimize the reward and hinder the protagonist.}
    \label{fig:pipeline}
\end{figure}
Adversarial training allows the protagonist to meet more severe situations, especially those dangerous but unexplored situations, thereby making it more adaptable to noisy environments. It also enables the adversary to learn to hinder the protagonist more effectively. Both the protagonist and the adversary receive feedback from the environment, and they both give an action based on the observation. The difference is that the adversary makes an action that tries to fail the task and makes the protagonist get a lower reward~\cite{gu2019adversary}. 
To simulate the real-world situation that the protagonist is disturbed by internal noise and external disturbance, we utilize two types of disturbances.

\paragraph{Internal robot joint disturbance}
Considering the actual robot with motor noise, the noisy action can be added into the robot joint space, which often happens when the joints motors of a robot are in bad condition. Formally, we will construct the noisy action as:
\begin{align}
\label{linear_add}
a = \hat{a} + \alpha \cdot \bar{a}, 
\end{align}
where $\hat{a}$ and $\bar{a}$ are the protagonist and adversary action of which each dimension is within the range [-1, 1], $\alpha$ is the parameter to control the maximum possible amplitude of the disturbance. For convenience, we will call $\alpha$ the amplitude parameter in the following text and figures.

\paragraph{External robot disturbance}
we use another robot acting to hinder the protagonist robot during the task, which is a perfect simulation when some human being or robot is really trying to disturb the protagonist robot and fail the task. Formally, we will construct the noisy action as:
\begin{align}
\label{concatenation_add}
a = ||[\hat{a}, \ \bar{a}], 
\end{align}
where $||$ represents concatenation, $\hat{a}$ is our protagonist robot action and $\bar{a}$ is our external adversarial robot action. We do not use a similar $\alpha$ in a) to constrain the $\bar{a}$ here, because we think it makes sense that the maximum possible amplitude of the disturbance on robot joint space can vary in reality, but it is not that reasonable to modify the action amplitude of another robot.

\begin{algorithm}[t!]
\caption{Adversarial training}
\begin{algorithmic}\label{alg:main}
\REQUIRE Randomly initialize deterministic protagonist actor $\mu_p$ and adversary actor $\mu_a$ with are parameterized as $\theta^{\mu_p}$, $\theta^{\mu_a}$,
randomly initialize protagonist critic network $Q_{\theta}(s,\hat{a})$ and adversary critic network $Q_{\phi}(s,\bar{a})$.
\FOR {iteration \textit{i} = $1$ to $N_{iter}$}
\STATE \textcolor{blue}{$\backslash\backslash$ update protagonist agent policy $\theta^{\mu_p}_i \gets \theta^{\mu_p}_{i-1}$.}
\FOR {episode=$1$ to $M$}
\STATE Sample $\hat{a}, \bar{a}$ from $\theta_{i-1}^{\mu_p}, \theta_{i-1}^{\mu_a}$. And execute action $a=\hat{a} + \alpha \bar{a}$ or $a=||[\hat{a}, \ \bar{a}]$ 
\STATE Observe reward $r$ and next state $s'$ from state $s$.
\STATE Relabel data $(s, \hat{a}, r, s')$ via HER.
\STATE Update protagonist critic $Q_{\theta}(s,\hat{a})$from Equation~\ref{td_error_p}.
\STATE Update protagonist actor $\theta^{\mu_p}$ from Equation~\ref{mu_p}.
\ENDFOR \\
\STATE \textcolor{blue}{$\backslash\backslash$ update adversary agent policy $\theta^{\mu_a}_i \gets \theta^{\mu_a}_{i-1}$.}
\FOR {episode=$1$ to $N$}
\STATE Sample $\hat{a}, \bar{a}$ from $\theta_i^{\mu_p}, \theta_{i-1}^{\mu_a}$. And execute action $a=\hat{a} + \alpha \bar{a}$ or $a=||[\hat{a}, \ \bar{a}]$ 
\STATE Observe reward $r$ and next state $s'$ from state $s$.
\STATE Relabel data $(s, \bar{a}, r, s')$ by HER.
\STATE Update adversary critic $Q_{\phi}(s,\bar{a})$ from Equation~\ref{td_error_a}.
\STATE Update adversary actor $\theta^{\mu_a}$ from Equation~\ref{mu_a}. 
\ENDFOR \\
\ENDFOR \\
\RETURN $\theta^{\mu_p}_{N_{iter}}$, $\theta^{\mu_a}_{N_{iter}}$
\end{algorithmic}
\end{algorithm}
\subsection{Adversary optimization}
We further analyze the practical optimization process for our adversarial training for the stochastic policies built on the soft actor-critic framework.

In the case of stochastic policy, we extend the standard soft actor-critic approach to our adversarial version, called \textit{Adv-SAC}.
Here we note the protagonist agent's actor and critic networks as $Q_p$ and $\mu_p$ and use $Q'_p$ and $\mu'_p$ to represent its target actor and critic networks. For the adversary agent, $Q_a$, $\mu_a$, $Q'_a$, $\mu'_a$ represent adversary actor, critic networks, and corresponding target actor and critic networks. 

As an entropy-regularized reinforcement learning, the critic objectives for both protagonist and adversary agents can be formulated as:
\begin{align}
\label{td_error_p}
L(\theta_{Q_p}) = \frac{1}{N}\sum_i{(y^p_i-Q_p(s_i, a_i|\theta^{Q_p}))^2},
\end{align}
\begin{align}
\label{td_error_a}
L(\theta_{Q_a}) = \frac{1}{N}\sum_i{(y^a_i-Q_a(s_i, a_i|\theta^{Q_a}))^2}, 
\end{align}
\noindent
where the protagonist target value:

$y^p_i=r_i+\gamma Q'_p(s_{i+1}, \mu_p(\cdot|s_{i+1})- \alpha \log \mu_p(\cdot|s_{i+1}))$.

\noindent
and the adversary target value:

$y^a_i=r_i+\gamma Q'_a(s_{i+1}, \mu_a(\cdot|s_{i+1})- \alpha \log \mu_a(\cdot|s_{i+1}))$.

We can note that the adversary agent critic network is the same with the protagonist where we just replace $Q_p$ $Q'_p$ and $\mu_p$ in Equation~\ref{td_error_p} with $Q_a$ $Q'_a$ and $\mu_a$. 

The main difference is from the policy optimization part. For the protagonist policy, our objective is to maximize the expected reward with maximizing entropy-regularized term.
\begin{equation}
\begin{split}
\label{mu_p}
  \nabla_{\theta^{\mu_p}} J(\theta) = &\nabla_a Q_p(s,a)\nabla_{\theta^{\mu_p}} \mu_p(s|\theta^{\mu_p}) \\
&+ \nabla_\theta \log \pi_\theta^{\mu_p}(a|s)Q_p^H(s, a).
\end{split}
\end{equation}

For the adversary policy, our objective is to minimize the expected reward with maximizing entropy-regularized term.
\begin{equation}
\begin{split}
\label{mu_a}
\nabla_{\theta^{\mu_a}} J(\theta)= &-\nabla_a Q_a(s,a)\nabla_{\theta^{\mu_a}} \mu_a(s|\theta^{\mu_a}) \\
&+ \nabla_\theta \log \pi_\theta^{\mu_a}(a|s)Q_a^H(s, a).
\end{split}
\end{equation}
The overall process is shown in the Figure~\ref{fig:pipeline} and the pseudocode are summarized in Alg~\autoref{alg:main}.

\section{Experiments}

For the experiments below, we investigate the following questions:
\begin{enumerate}
    \item Under the disturbance from internal robot joints and external environment, is the protagonist manipulation policy robust enough to accomplish the given task?
    \item Is our adversarial attack a more aggressive disturbance than random noise in the robot manipulation tasks?
    \item Can the results of the real-world experiment support a similar conclusion with the simulation experiment results? 
\end{enumerate}

We conduct 1) robot claw turning task \textit{DClawTurnFixed} on ROBEL platform\cite{ahn2020robel} in simulation environment and real-world environment; 2) the \textit{DoubleArmPick} task custom in the gym~\cite{openai_gym} environment. In the first task, we add internal disturbance on Robel robot joints. In the second task, we use another adversarial robot to execute external disturbance on the protagonist robot.

\begin{figure}[t!]
    \centering
    \includegraphics[width=0.45\textwidth]{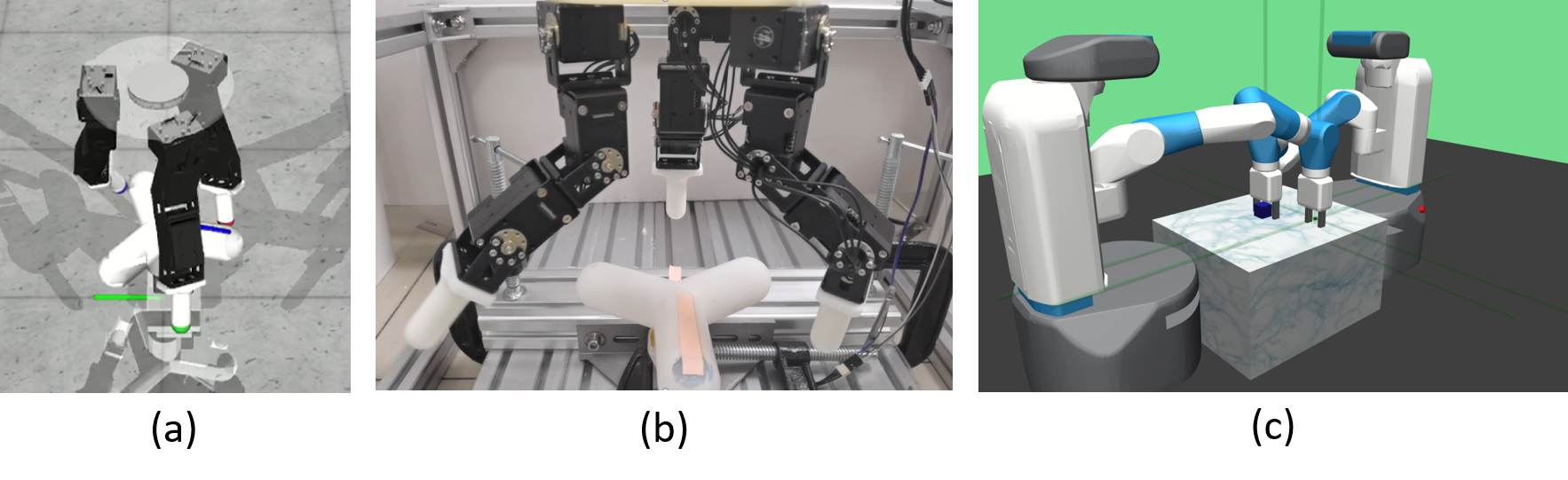}
    \caption{
    (a) \textit{DClawTurnFixed} task in ROBEL\cite{ahn2020robel} simulation environment. (b) Real-world \textit{DClawTurnFixed} platform using ROBEL\cite{ahn2020robel} D'Claw robot. (c) \textit{DoubleArmPick} task custom in gym~\cite{openai_gym} environment.}
    \label{fig:platform}
\end{figure}

\begin{figure*}
    \centering
    \includegraphics[width=.72\textwidth]{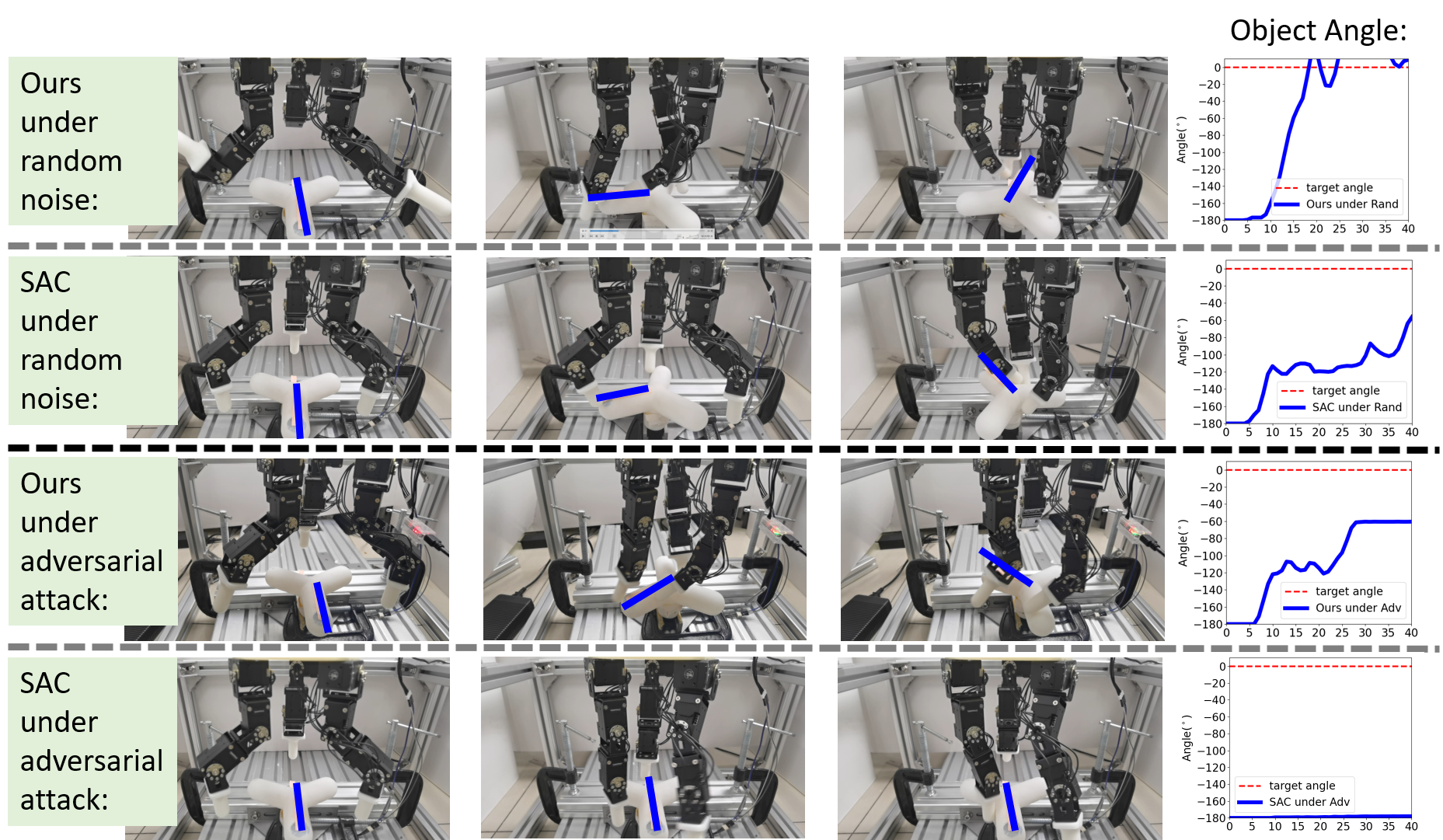}
    \caption{We mark a blue line on the triple-arm object to indicate its current angle. The last column of curves shows how the object angle varies with the time steps in a game. Once the blue curve rises across the target angle marked by the dotted red line, the robot successes. The adversarial attack or random noise with an amplitude 0.8 is added to the robot joint space. The first row of figures shows that our protagonist can easily overcome the random noise and accomplish the task. The second row of figures shows that the normal SAC agent can turn the object for about 150$^\circ$ under the random noise. The third row of figures shows that our protagonist can turn the object for about 120$^\circ$ under the adversarial attack. The fourth row of figures shows that the normal SAC agent is totally out of function under the adversarial attack.}
    \label{fig:real-robel}
\end{figure*}

The \textit{DClawTurnFixed} task in the ROBEL environment requires the robot claw(D'Claw) to turn the triple-arm object over 180$^\circ$. The triple-arm object is initialized at a fixed angle -180$^\circ$. If the robot claw turns the object to above 0 $^\circ$ once within 40 time steps, it successes in this game. As is shown in Fig.\ref{fig:platform} (b), in real-world experiments, we use orange markers on the object and the ground to mark the object angle and the target angle. In Fig.\ref{fig:real-robel}, we use a blue line to show the object angle clearer. The \textit{DoubleArmPick} task is a double robot version of the \textit{FetchPickAndPick} task in the OpenAI gym~\cite{openai_gym}. It requires a robot to pick the block on the table and place it at the goal position in the air or on the table within 50 time steps. Both block and goal positions are initialized randomly. The first and second tasks are set in dense and sparse reward modes separately, indicating our method is suitable for both situations.
\begin{figure}[t!]
    \centering
    \includegraphics[width=0.5\textwidth]{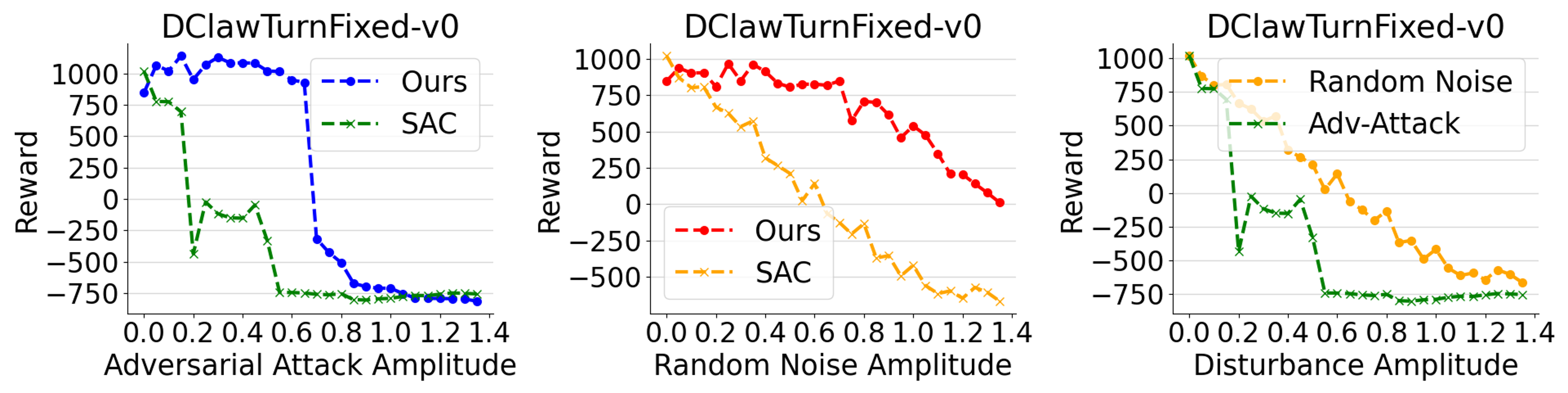}
    \caption{(a) Compared with normal SAC agent, our protagonist gets obviously higher reward against adversarial attack on joint space. (b) Compared with normal SAC agent, our protagonist gets higher reward against random noise on joint space. (c) Attacking the same normal SAC agent on its joint space, the adversary decreases the reward more dramatically than random noise. }
    \label{fig:robel_simulation_test}
\end{figure}

\subsection{Internal Disturbance on Robel Robot} 
In the Robel robot experiments, we simulate the situation when the joint motors of a robot are in bad condition and have noise on them. We hope to figure out how the noise on the joint motors will influence the robot performance and develop a robust policy for robots to overcome the disturbance on joint space. In this case, the ideal robot action on joint space is represented as $\hat{a}$. The disturbance on joint space is represented as $\alpha \bar{a}$, where $\alpha$ stands for the maximum possible amplitude of the disturbance, and $\bar{a}$ represents its character, which can be random noise or the output action of an adversarial agent. Each dimension of $\hat{a}$ and $\bar{a}$ is within the range [-1, 1]. The random noise obeys the uniform distribution on [-1, 1]. So the real action on the robot joints can be given by $a=\hat{a} + \alpha \bar{a}$, which means the real action on each joint is the linear combination of the ideal action and a disturbance with a maximum possible amplitude $\alpha$. We will call $\alpha$ amplitude in short in the following text and figures.

\paragraph{Simulation Experiment}
As is shown in Fig.\ref{fig:platform}(a) (b), a Robel D’Claw robot is trained to accomplish the \textit{DClawTurnFixed} task. 

Utilizing the adversarial SAC training method, a protagonist policy and an adversary policy is trained alternately, and each is trained for 20,000 episodes. A normal SAC policy is also trained for 20,000 episodes to be tested as a contrast baseline. We choose $\alpha=0.6$ during the adversarial training process because after lots of experiments, we find that this $\alpha$ value can help train the most robust protagonist and the most aggressive adversary in this task. Then we test the trained policies in a wide range of $\alpha$ value to test the adaptability of the trained policies.

We use the output action $\bar{a}$ of the adversary to attack the protagonist and the normal SAC agent. $\alpha$ varies from 0.0 to 1.4 and the gained rewards of both agents are shown in Fig.\ref{fig:robel_simulation_test} (a). When $\alpha$ is below 0.2, both agents can accomplish the task and get a high reward. When $\alpha$ is between 0.2 and 0.6, our protagonist can still accomplish the task perfectly and maintain the high reward, while the normal SAC agent fails and gets a meager reward. When $\alpha$ rises to above 0.8, both agents fail.

Fig\ref{fig:robel_simulation_test}(b) suggests that our protagonist has higher robustness than the normal SAC agent against the random noise on joint space. When the amplitude $\alpha$ of the random noise rises, the rewards of both agents will decline, but the protagonist can keep getting an higher reward than the SAC agent in a large range of $\alpha$, indicating that the protagonist accomplishes the task better in the same noisy condition.

The effect of adversarial attack and random noise on normal SAC agent are compared in Fig.\ref{fig:robel_simulation_test} (c). At any same amplitude $\alpha$ below 1.4, the reward under adversarial attack is significantly lower than that under random noise, showing that adversarial attack is smarter and more aggressive than the random noise to hinder the protagonist and fail the task.

To sum up, the simulation experiments on the Robel robot show that the trained protagonist can obtain significantly better robustness against the adversarial attack and random noise. Moreover, the trained adversary appears to be far more aggressive than a random attack in all these experiments.

\paragraph{Real-World Experiment}
We also conducted real-world experiments on the Robel D’Claw robot in the same task. As is shown in Fig.\ref{fig:real-robel}, the D’Claw robot is hung on a frame, and the triple-arm object with a motor on its base is fixed on the floor to record the angle of the object during the experiments and reset the object to the -180$^\circ$ initial position before each game. 

Since the simulation environment and our real-world experiment environment are basically consistent, we apply the protagonist policy and the adversarial policy trained in the simulation environment directly on the real robot. Also, a normal SAC policy is trained in the simulation environment and used on the real robot for contrast. Fig.\ref{fig:real-robel} shows the situation when the amplitudes of adversarial attack or random noise on the joint space are both 0.8. The first three columns of Fig.\ref{fig:real-robel} shows the process of turning the object. The last column includes the curves of the object angle varying with the time step. In each game, if the robot claw can turn the object form -180 $^\circ$ to over 0 $^\circ$, it successes.

In the first row of Fig.\ref{fig:real-robel}, when the protagonist is acting against a random noise with amplitude 0.8, it can easily accomplish the task. The curve indicates that the object is turned over 180 $^\circ$ very quickly and table \ref{tab:single_grasp} shows a success rate around 94\%. In contrast, when the normal SAC agent acts against random noise, it sometimes fails to turn the object to an angle large enough to succeed, getting a success rate of around 82\%. Its turning process and the object angle curve is recorded in the second row of Fig.\ref{fig:real-robel}. The lower two rows of Fig.\ref{fig:real-robel} compare the performance of the protagonist and normal sac agent against adversarial attack. Our protagonist can turn the object to about 120 $^\circ$ but rarely over 180 $^\circ$ to success. On the contrary, the normal SAC agent becomes totally out of function and can hardly move the object at all.

Considering that the real world platform is impossible to be in the exact same condition as the simulation platform, the results of the real-world experiment are slightly different from the results of its simulation experiments on some specific numbers, but the general trends are the same. Both real-world and simulation results indicate that the adversarial training can train a protagonist with higher robustness and an adversary, which is more aggressive than random noise. More details can be seen in the videos.

\subsection{External Disturbance in \textit{DoubleArmPick} Task} 

\begin{figure}[t!]
    \centering
    \includegraphics[width=0.5\textwidth]{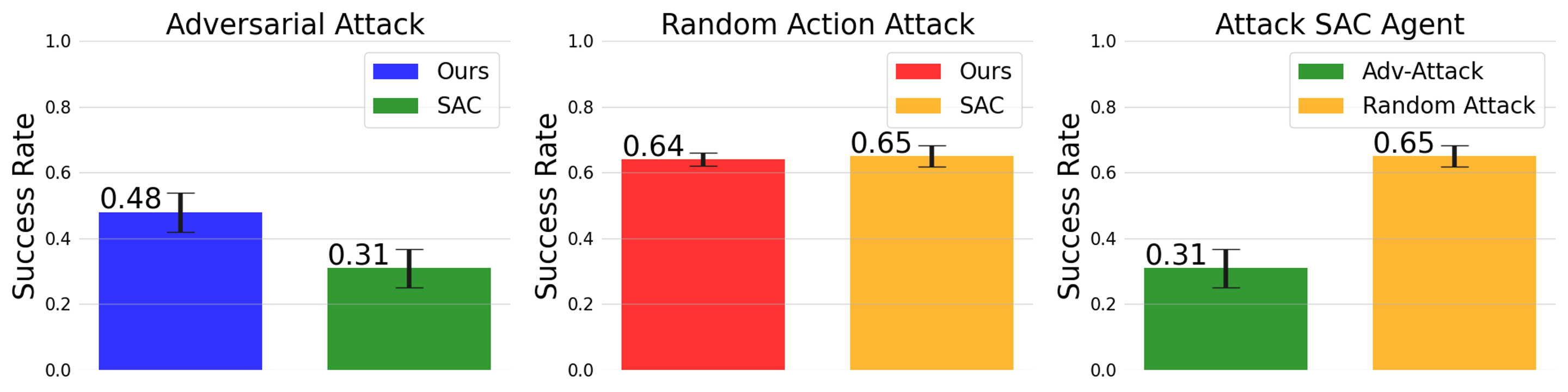}
    \caption{In \textit{DoubleArmPick} task: (a) Under the adversarial attack of another robot, our protagonist can get about 20\% higher success rate than the normal SAC agent. (b) When the adversarial robot takes random action, our protagonist and normal SAC agent have similar performance in success rate. (c) Adversarial attack from another robot is far more aggressive than a random attack, causing the normal SAC agent an extra 30\% drop in success rate. }
    \label{fig:double arm bar}
\end{figure}

Besides the internal disturbance like motor noise, which can be described as $\alpha \bar{a}$ added on the joint space of a robot, there are also some cases when the robot is disturbed by a real person or another real robot. To simulate this kind of situation, we build a double robot environment in gym~\cite{openai_gym}, as shown in Fig.\ref{fig:platform} (c). The robot on the right side of the table is the protagonist who tries to accomplish the \textit{DoubleArmPick} task in sparse reward mode, while the robot on the left side is the adversary who tries to hinder the protagonist and fail the task.

Using our proposed adversarial training method based on SAC, we train a protagonist for the right robot and an adversary for the left robot. Also, we lock the left robot (setting its action to zero in every step) and train a normal SAC agent on the right robot for the \textit{DoubleArmPick} task as a contrast baseline for the following tests.

The protagonist or normal SAC agent on the right is tested in the \textit{DoubleArmPick} environment against the adversary robot on the left. Fig.\ref{fig:double arm bar} (a) shows the experimental results when the left robot executes the trained adversarial policy. Fighting against the adversarial robot, the protagonist can get about 50\% of success rate in the \textit{DoubleArmPick} task, while the normal SAC agent can only reach around 30\% of success rate, indicating that the adversarial training can improve the robustness of the protagonist agent against the disturbance of another robot.

When the adversarial robot on the left takes a random action instead of the output from an adversarial policy, the adversarial trained protagonist can maintain a performance close to the normal SAC agent. As shown in Fig.\ref{fig:double arm bar} (b), it can reach a success rate of above 64\%, slightly lower than that of the normal SAC agent.

Compare the success rate of the normal SAC agent against adversarial action and random action in Fig.\ref{fig:double arm bar} (c), it is evident that the trained adversarial policy can take a far more aggressive attack than the random action, causing an extra 30\% decline in the success rate of the task. The mean success rate in table \ref{tab:single_grasp} of each setting is calculated after 50 times repeated experiments.

\begin{figure}[t!]
    \centering
    \includegraphics[width=0.5\textwidth]{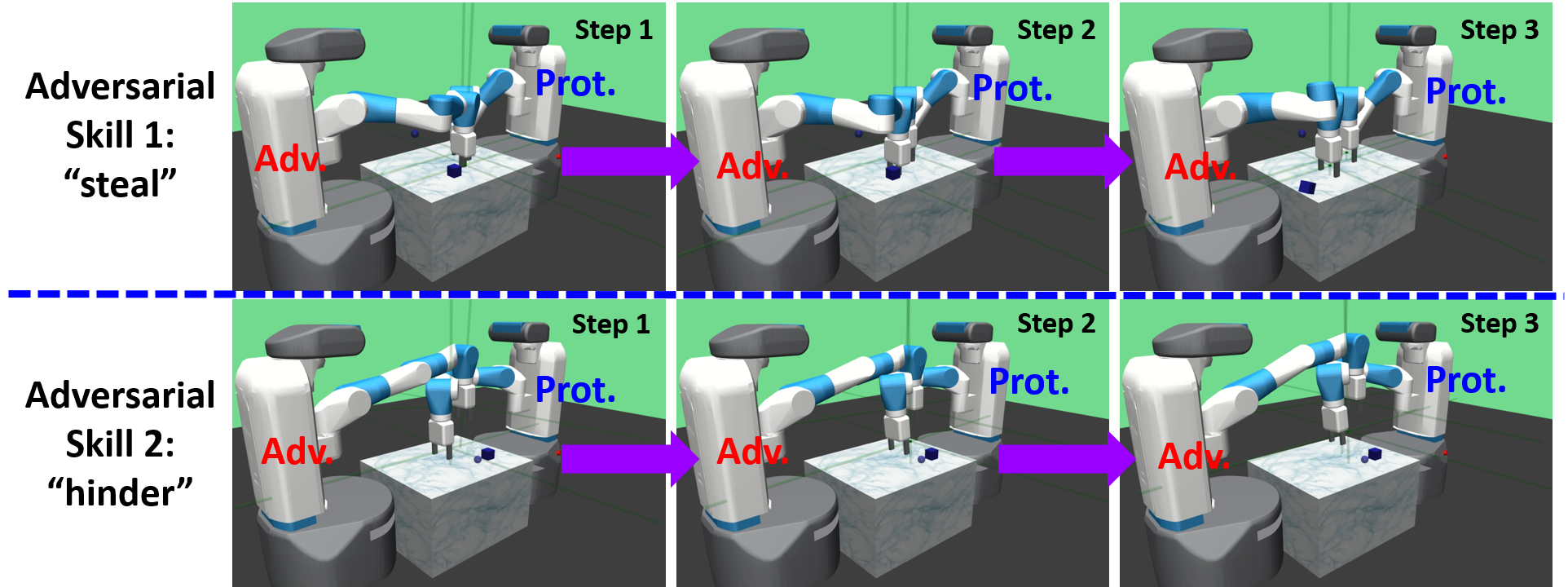}
    \caption{The first row of three figures shows the adversarial robot's process on the left push the block off the table to fail the task. We call this skill "stealing the object". The second row of the three figures shows the adversarial robot on the left put its arm between the protagonist and the block, preventing the protagonist from getting the object. We call this skill "hindering the protagonist".}
    \label{fig:double-arm skills}
\end{figure}

Fig.\ref{fig:double-arm skills} shows two of the typical skills learned by the adversary to fail the task. In the first row of Fig.\ref{fig:double-arm skills}, the block is initialized close to the adversarial robot. The adversary will reach it ahead of the protagonist and then push it off the table, making the protagonist impossible to accomplish the task. In the second row of Fig.\ref{fig:double-arm skills}, the block is initialized close to the protagonist. In this case, it is hard for the adversary to reach the block first and push it off the table, so it will stretch its arm far ahead and put it in the way of the protagonist robot arm to the block. By doing so, it can hinder the protagonist from reaching the block and thereby fail the task. There are also many other interesting skills learned by the protagonist and the adversary. For example, the two agents will compete to get the block in some cases. Sometimes, the adversary will push away the block that has already been placed in the protagonist's goal position. Sometimes, the protagonist will push away the adversary gripper to get the block. Please find them in the video. 

\begin{table}[ht]
\centering
\caption{success rate of 'DClawTurnFixed' experiments}
\begin{tabular}{cccccc}
\hlineB{2}
Method & Disturbance & Amp=0.0 & Amp=0.5 & Amp=0.8 & Amp=1.0  \\ \hline
Ours & Rand. & 100\% & 98\% & 94\% & 92\% \\ \hline
SAC & Rand. & 100\% & 88\% & 82\% & 38\% \\ \hline
Ours & Adv. & 100\% & 86\% & 4\% & 2\% \\ \hline
SAC & Adv. & 100\% & 2\% & 0\% & 0\% \\
\hlineB{2}
\end{tabular}
\label{tab:single_grasp}
\end{table}

\section{Conclusion}
In this paper, we propose an adversarial training algorithm for robotic manipulation tasks and carry out abundant experiments to test the robustness of the trained protagonist and the disturbance ability of the adversary. This algorithm can improve the policy robustness against both the internal disturbance in robot joint space and the external disturbance from another robot. We believe our algorithm can be widely useful in real-world robot applications against inevitable internal and external disturbances. Besides, to the best of our knowledge, this is also the first time that researchers successfully apply the adversarial training method on sparse reward RL tasks. According to the results of both simulation and real-world experiments, our trained protagonist is more robust than normal RL policy, and the trained adversary is obviously more aggressive than random disturbance. More real-world experiments on other robotic manipulation tasks can be a good direction for future work.

\clearpage
\bibliographystyle{IEEEtran} 
\bibliography{IEEEabrv,ref}

\end{document}